# Expert System Based On Neural-Fuzzy Rules for Thyroid Diseases Diagnosis

**Ahmad Taher Azar[1,3], IEEE Senior Member, Aboul Ella Hassanien[2,3]**

[1]Faculty of Computers and Information Benha University, Egypt.

[1]Faculty of Computers and Information - Cairo University
[3]Scientific Research Group in Egypt (SRGE) http://www.egyptscience.net

{Ahmad_t_azar@ieee.org; aboitcairo@gmail.com}

**Abstract.** The thyroid, an endocrine gland that secretes hormones in the blood, circulates its products to all tissues of the body, where they control vital functions in every cell. Normal levels of thyroid hormone help the brain, heart, intestines, muscles and reproductive system function normally. Thyroid hormones control the metabolism of the body. Abnormalities of thyroid function are usually related to production of too little thyroid hormone (hypothyroidism) or production of too much thyroid hormone (hyperthyroidism). Therefore, the correct diagnosis of these diseases is very important topic. In this study, Linguistic Hedges Neural-Fuzzy Classifier with Selected Features (LHNFCSF) is presented for diagnosis of thyroid diseases. The performance evaluation of this system is estimated by using classification accuracy and k-fold cross-validation. The results indicated that the classification accuracy without feature selection was 98.6047% and 97.6744% during training and testing phases, respectively with RMSE of 0.02335. After applying feature selection algorithm, LHNFCSF achieved 100% for all cluster sizes during training phase. However, in the testing phase LHNFCSF achieved 88.3721% using one cluster for each class, 90.6977% using two clusters, 91.8605% using three clusters and 97.6744% using four clusters for each class and 12 fuzzy rules. The obtained classification accuracy was very promising with regard to the other classification applications in literature for this problem.

**Keywords:** Thyroid Disorders; Soft Computing; Takagi-Sugeno-Kang (TSK) fuzzy inference system; Linguistic Hedge (LH); Feature selection (FS).

## 1 Introduction

The thyroid gland is the biggest gland in the neck [1]. The thyroid gland is placed in the anterior neck. It has the shape of a butterfly with the two wings being represented by the left and right thyroid lobes. The thyroid provides the thyroid hormones. The thyroid gland produces two active thyroid hormones, levothyroxine (abbreviated T4) and triiodothyronine (abbreviated T3). These hormones are important in the production of proteins, in the regulation of body temperature, and in overall energy production and regulation. The thyroid gland has many diseases. The most of these are goiters, thyroid cancer, solitary thyroid nodules, hyperthyroidism, hypothyroidism, thyroiditis, etc [2]. The hypothyroidism is too little thyroid hormone. It is a common problem. Hypothyroidism can even be associated with pregnancy. The diagnosis and treatment for all types of hypothyroidism is usually straightforward. The goiter is a dramatic enlargement of the thyroid gland. Goiters are often removed due to cosmetic reasons. Moreover, these compress other vital structures of the neck including the trachea and the esophagus making breathing and swallowing difficult. The thyroid nodules can take on characteristics of malignancy. Therefore, these require biopsy or surgical excision. In addition to, these contain risks of radiation exposure. The thyroid cancer is a fairly common malignancy. Therefore the diagnosis of this disease is very difficult. The hyperthyroidism is too much thyroid hormone. The radioactive iodine, anti-thyroid drugs, or surgery are common methods used for treating a hyperthyroid patient. The thyroiditis is an inflammatory status for the thyroid gland. This can give with a number of symptoms such as fever and pain, but it can also give as subtle findings of hypo or hyperthyroidism. Accurate prediction of the thyroid data besides clinical examination and complementary investigation is an important issue in the diagnosis of thyroid disease. Various new methods have been used for diagnosis of thyroid diseases like Artificial Neural Network [3-8], Linear Discriminant Analysis (LDA) [9], decision trees [9, 10], Fuzzy expert systems and neuro-fuzzy classification [11-16], Support Vector Machines [15-20]. This paper presents a fuzzy feature selection (FS) method based on the linguistic hedges (LH) concept [21, 22] for thyroid diseases classification. This classifier is used to achieve a very fast, simple and efficient computer aided diagnosis (CAD) system. The rest of this paper is organized as follows: Section 2 provides subjects and methods that are used in this study. In Section 3, a review of the classifier that is considered in thyroid diseases diagnosis is presented. Section 4 reports the results of experimental evaluations of the adaptive neural-fuzzy classifier and finally, in Section 5, conclusion and directions for future research are presented.

## 2  Subjects and Methods

Classification of data from the University of California, Irvine (UCI) machine learning data set repository was performed to evaluate the effectiveness of the Neural-fuzzy classifier on real-world data, and to facilitate comparison with other classifiers. [23]. The dataset contains 3 classes and 215 samples. These classes are assigned to the values that correspond to the hyper-, hypo-, and normal function of the thyroid gland. All samples have five features. These are [23]:

 1. T3-resin uptake test (A percentage).
 2. Total serum thyroxin as measured by the isotopic displacement method.
 3. Total serum triiodothyronine as measured by radioimmuno assay.
 4. Basal thyroid-stimulating hormone (TSH) as measured by radioimmuno assay.
 5. Maximal absolute difference of TSH value after injection of 200 µg of thyrotropin-releasing hormone as compared to the basal value.

The 150 samples of 215 belong to hyper-function class namely class-1. The 35 samples of 215 belong to hypo-function class namely class-2. The 30 samples of 215 belong to normal-function class namely class-3 [23].

## 3. Adaptive Neuro-Fuzzy Classifiers

The usage of ANFIS [24-26] for classifications is unfavorable. For example, if there are three classes labeled as 1, 2 and 3. The ANFIS outputs are not integer. For that reason the ANFIS outputs are rounded, and determined the class labels. But, sometimes, ANFIS can give 0 or 4 class labels. These situations are not accepted. As a result ANFIS is not suitable for classification problems. In this section, adaptive neuro-fuzzy classifier is discussed in details. In these models, k-means algorithm is used to initialize the fuzzy rules. Also, Gaussian membership function is only used for fuzzy set descriptions, because of its simple derivative expressions.

### 3.1 Adaptive Neuro-Fuzzy Classifier with Linguistic Hedges (ANFCLH)

Adaptive neuro-fuzzy classifier (ANFC) with Linguistic hedges [21] is based on fuzzy rules. Linguistic hedges are applied to the fuzzy sets of rules, and are adapted by Scaled Conjugate Gradient (SCG) algorithm. By this way, some distinctive features are emphasized by power values, and some irrelevant features are damped with power values. The power effects in any feature are generally different for different classes. The using of linguistic hedges increases the recognition rates. A fuzzy classification rule that has two inputs $\{x_1, x_2\}$ and one output $y$ is defined with LHs as IF $x_1$ is $A_1$ with $p_1$ hedge AND $x_2$ is $A_2$ with $p_2$ hedge THEN $y$ is $C_1$ class, where $A_1$ and $A_2$ denote linguistic terms that are defined on $X_1$ and $X_2$ feature space; $p_1$ and $p_2$ denote linguistic hedges, respectively; $C_1$ denotes the class label of the output y. Fig. 1 shows the ANFCLH architecture. The feature space with two inputs $\{x_1, x_2\}$ is partitioned into three classes $\{C_1, C_2, C_3\}$, in the Figure. The feature space $X_1 \times X_2$ is separated into fuzzy regions (Jang et al. 1997). This technique is based on zero-order Sugeno fuzzy model [27]. The crisp outputs of fuzzy rules are determined by weighted average operator [26].

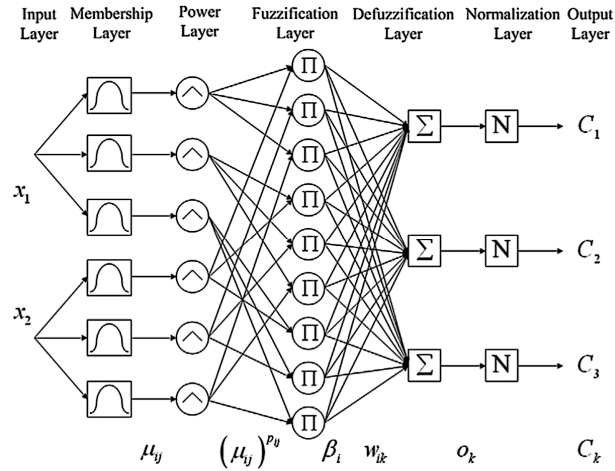

Fig. 1 A neuro-fuzzy classifier with LHs. [21]

In this classifier, the nodes in the same layer have the same type of node functions. The layers and their properties are given as follows:

*Layer 1*: In this layer, the membership grade of each input to specified fuzzy region is measured. Gaussian function is employed as MF due to smooth partial derivatives of its parameters, and has less parameter. The Gaussian MF is given as follows [21]:

$$\mu_{ij}(x_{sj}) = \exp\left(-0.5\frac{(x_{sj}-c_{ij})^2}{\sigma_{ij}^2}\right) \quad (1)$$

where $\mu_{ij}(x_{sj})$ represents the membership grade of the $i$th rule and the $j$th feature; $x_{sj}$ denotes the $s$th sample and the $j$th feature of input matrix $X\{X \in R^{NxD}\}$; $c_{ij}$ and $\sigma_{ij}$ are the center and the width of Gaussian function, respectively.

*Layer 2*: In this layer, the secondary meanings of fuzzy sets are calculated with their LHs as in Eq. (2) [21].

$$\alpha_{ijs} = [\mu_{ij}(x_{sj})]^{p_{ij}} \quad (2)$$

where $\alpha_{ijs}$ denotes the modified membership grades of $\mu_{ij}(x_{sj})$; $p_{ij}$ denotes the LH value of the $i$th rule and the $j$th feature.

*Layer 2*: The degree of fulfillment of the fuzzy rule for x s sample is determined in this layer. It is also called as the firing strength of rule. So, the $B_{is}$ firing strength of the $i$th rule for D number of features is defined as in Eq. (3) [21].

$$B_{is} = \prod_{j=1}^{D} \alpha_{ijs} \quad (3)$$

*Layer 4*: In this layer, the weighted outputs are calculated as in Eq. (4) [21], and every rule can affect each class according to their weights. However, if a rule controls a specific class region, the weight between this rule output and the specific class is to be bigger than the other class weights. Otherwise, the class weights are fairly small:

$$O_{sk} = \sum_{i=1}^{U} \beta_{is} w_{ik} \quad (4)$$

where $w_{ik}$ represents the degree of belonging to the $k$th class that is controlled with the $i$th rule; $O_{sk}$ denotes the weighted output for the $s$th sample that belongs to the $k$th class, and $U$ is the number of rules.

*Layer 5*: If the summation of weights is bigger than 1, the outputs of the network should be normalized in the last layer as follows [21]:

$$h_{sk} = \frac{O_{sk}}{\sum_{l=1}^{k} O_{sl}} = \frac{O_{sk}}{\delta_s}, \delta_s = \sum_{l=1}^{k} O_{sl} \quad (5)$$

Where $h_{sk}$ represents the normalized degree of the $s$th sample that belongs to the $k$th class; and $K$ is the number of classes. After then, the class label ($C_s$) of $s$th sample is determined by the maximum $h_{sk}$ value as in Eq. (6) [21].

$$C_s = \max_{k=1,2,\ldots,k} \{h_{sk}\} \qquad (6)$$

The antecedent parameters of the network {$c, \sigma, p$} could be adapted by any optimization method. In this study, scaled conjugate gradient (SCG) method is used to adapt the network parameters [28]. The cost function that is used in the SCG method is determined from the least mean squares of the difference target and the calculated class value [26, 29]. According to the above definition, the cost function $E$ is defined as in Eq. (7) [21].

$$E = \frac{1}{N}\sum_{s=1}^{N} E_s, \quad E_s = \frac{1}{2}\sum_{k=1}^{K}(t_{sk} - h_{sk})^2 \qquad (7)$$

### 3.2 Linguistic Hedges Neural-Fuzzy Classifier With Selected Features (LHNFCSF)

Cetişli (2010) presented a fuzzy feature selection (FS) method based on the LH concept. It uses the powers of fuzzy sets for feature selection [21, 22]. The values of LHs can be used to show the importance degree of fuzzy sets. When this property is used for classification problems, and every class is defined by a fuzzy classification rule, the LHs of every fuzzy set denote the importance degree of input features. If the LHs values of features are close to concentration values, these features are more important or relevant, and can be selected. On the contrary, if the LH values of features are close to dilation values, these features are not important, and can be eliminated. According to the LHs value of features, the redundant, noisily features can be eliminated, and significant features can be selected. In this technique, [22], if linguistic hedge values of classes in any feature are bigger than 0.5 and close to 1, this feature is relevant, otherwise it is irrelevant. The program creates a feature selection and a rejection criterion by using power values of features. There are two selection criteria, one is the selection of features that have the biggest hedge value for any class and the other is the selection of features that have a bigger hedge value for every class, because any feature cannot be selective for every class. For that reason, a selective function should be described from the hedge values of any feature as in Eq. (8) [22]:

$$p_j = \prod_{i=1}^{k} p_{ij} \qquad (8)$$

where $P_j$ denotes the selection value of the jth feature, and K is the number of classes. The Feature selection and classification algorithms were discussed in detail in [22].

## 4. Results and Discussions

**4.1 Training and Testing Phases of Classifier**

The collection of well-distributed, sufficient, and accurately measured input data is the basic requirement in order to obtain an accurate model. The classification process starts by obtaining a data set (input-output data pairs) and dividing it into a training set and testing data set. The training data set is used to train the NFC, whereas the test data set is used to verify the accuracy and effectiveness of the trained NFC. Once the model structure and parameters have been identified, it is necessary to validate the quality of the resulting model. In principle, the model validation should not only validate the accuracy of the model, but also verify whether the model can be easily interpreted to give a better understanding of the modeled process. It is therefore important to combine data-driven validation, aiming at checking the accuracy and robustness of the model, with more subjective validation, concerning the interpretability of the model. There will usually be a challenge between flexibility and interpretability, the outcome of which will depend on their relative importance for a given application. While, it is evident that numerous cross-validation methods exist, the choice of the suitable cross-validation method to be employed in the NFC is based on a trade-off between maximizing method accuracy and stability and minimizing the operation time. To avoid overfitting problems during modeling process, k-fold cross-validation was used for better reliability of test results [30]. In k-fold cross-validation, the original sample is randomly partitioned into k subsamples. A single subsample is retained as the validation data for testing the model, and the remaining k - 1 subsamples are used as training data. The cross-validation process is then repeated k times (the 'folds'), with each of the k subsamples used exactly once as the validation data. The average of the k results gives the validation accuracy of the algorithm [31]. The advantages of k-fold cross validation are that the impact of data dependency is minimized and the reliability of the results can be improved [32]. In the first phase, NFC is trained using all data instances without feature reduction. In this study, 60–40% partition was used for training-test of the NFC for diagnosis of thyroid disease. According to this proportion, 129 of 215 samples in thyroid gland database were used for training of NFC while 86 of 215 samples in thyroid gland database were used for testing. The number of training and test data for each of classes can be given as in Table 1.

**Table 1.** The number of training and test data for each of class.

| Class | The number of training data (60%) | The number of testing data (40%) |
|---|---|---|
| Class-1: The hyper-function class | 90 | 60 |
| Class-2: The hypo-function class | 21 | 14 |
| Class-3: The normal-function class | 18 | 12 |
| Total | 129 | 86 |

The error convergence curve of NFC achieved mean RMSE values of 0.0233 in the training phase as shown in Fig. 2. In feature reduction stage of the ANFCLH for diagnosis of thyroid disease, the feature extraction and the feature reduction processes are performed. In the validation phase, 4-fold cross validation is used to compute the recognition rates. The number of fuzzy rules is determined according to the number of classes. The LH values of selected features for thyroid gland dataset after the training are given in Table 2. The total LH values for every class and every feature are shown in Fig. 3.

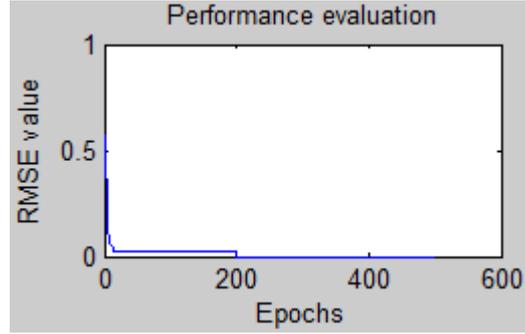

Fig. 2 Performance Evaluation of NFC during training phase without feature reduction

According to the feature selection algorithm, T3-resin uptake test (Feature 1), total serum thyroxin (Feature 2), Total serum triiodothyronine (Feature 3) and Maximal absolute difference of TSH (Feature 5) are common relevant features for each class. The basal thyroid-stimulating hormone (TSH) (Feature 4) is irrelevant for each class. It can be seen from Table 2 that malignant class is easily distinguished from the other class.

**Table 2**. The LH values of Thyroid Disease dataset for every class and every feature.

| Class/Features | Feature 1 | Feature 2 | Feature 3 | Feature 4 | Feature 5 |
|---|---|---|---|---|---|
| Class-1: The hyper-function class | 0.7147 | 0.5382 | 0.2665 | 5.68e-11 | 0.3119 |
| Class-2: The hypo-function class | 0.4052 | 1 | 0.3439 | 0.2792 | 0.8821 |
| Class-3: The normal-function class | 0.6538 | 1 | 0.4367 | 0.3828 | 0.5506 |
| Total LH values | 1.7738 | 2.5383 | 1.0471 | 0.6619 | 1.7446 |

If the classification rules are expressed for each class, then the rules are:

**R1:** IF T3-resin uptake test is $A_{11}$ with $p_{11} = 0.7147$ AND total serum thyroxin is $A_{12}$ with $p_{12} = 0.2665$ AND Total serum triiodothyronine is $A_{13}$ with $p_{13} = 0.2665$ AND Basal thyroid-stimulating hormone (TSH) is $A_{14}$ with $p_{14} = 5.6802e-11$ AND Maximal absolute difference of TSH is $A_{15}$ with $p_{15} = 0.3119$ THEN class is hyper-function

**R2:** IF T3-resin uptake test is $A_{21}$ with $p_{21} = 0.4052$ AND total serum thyroxin is $A_{22}$ with $p_{22} = 1$ AND Total serum triiodothyronine is $A_{23}$ with $p_{23} = 0.3439$ AND Basal thyroid-stimulating hormone (TSH) is $A_{24}$ with $p_{24} = 0.2792$ AND Maximal absolute difference of TSH is $A_{25}$ with $p_{25} = 0.8821$ THEN class is hypo-function.

**R3:** IF T3-resin uptake test is $A_{31}$ with $p_{31} = 0.6538$ AND total serum thyroxin is $A_{32}$ with $p_{32} = 1$ AND Total serum triiodothyronine is $A_{33}$ with $p_{33} = 0.4367$ AND Basal thyroid-stimulating hormone (TSH) is $A_{34}$ with $p_{34} = 0.3828$ AND Maximal absolute difference of TSH is $A_{35}$ with $p_{35} = 0.5506$ THEN class is normal-function.

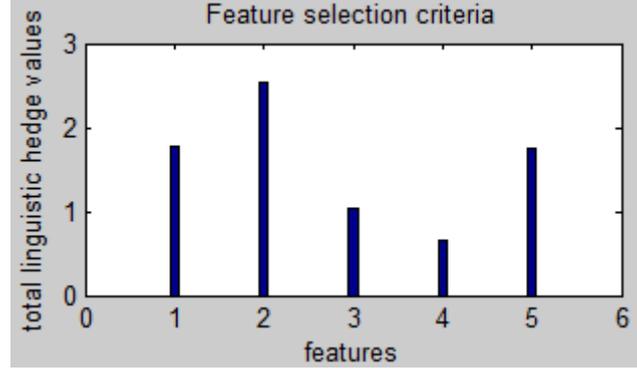

Fig. 3 Total LH values of thyroid gland dataset for every class and every feature.

After the classification step, it can be seen from Table 3, using one cluster for each class, that some of the hedge values are bigger than 1, because the hedge values are not constrained in the classification step [22]. It's clear from the results that T3-resin uptake test (Feature 1) is very important feature for hyper-function and normal-function classes while total serum thyroxin (Feature 2) is very important feature for hypo-function class. As shown in Table 3, the discriminative powers of the selected features are better than all features.

**Table 3.** The LH values of thyroid disease dataset for every class after selection of relevant features

| Class/Features | Feature 2 | Feature 1 | Feature 5 | Feature 3 |
|---|---|---|---|---|
| Class-1: The hyper-function class | 0.6966 | 1.0668 | 0.5294 | 1.0210 |
| Class-2: The hypo-function class | 1.3054 | 1.0393 | 1.0142 | 1.1147 |
| Class-3: The normal-function class | 1.3278 | 1.4968 | 1.1150 | 0.9845 |

The classification results of the training and testing phases obtained from the neural-fuzzy classifier are displayed in Table 4 and also represented graphically in Fig. 4. Here, each class for LHNFCSF is intuitively defined with 3, 6, 9 and 12 fuzzy rules based on the cluster size for each class ranged from 1-4 clusters. The results indicated that the classification accuracy without feature selection was 98.6047% and 97.6744% during training and testing phases, respectively with RMSE of 0.02335. After applying feature selection algorithm, LHNFCSF achieved 100% for all cluster sizes during training phase. However, in the testing phase LHNFCSF achieved 88.3721% using one cluster for each class, 90.6977% using two clusters, 91.8605% using three clusters and 97.6744% using four clus-ters for each class and 12 fuzzy rules. The classifier achieved mean RMSE values of 7.4547e-8 in the training phase using four clusters as shown in Fig. 5.

**Table 4**. The LHNFCSF classification results of thyroid disease dataset.

| Features | Cluster size for each class | Training Accuracy | Testing Accuracy | RMSE | No. of Rules |
|---|---|---|---|---|---|
| All | 1 | 98.6047 | 97.6744 | 0.02335 | 3 |
| 1, 2, 3, 5 | 1 | 100 | 88.3721 | 0.00091 | 3 |
| All | 2 | 98.6047 | 97.6744 | 0.02335 | 6 |
| 1, 2, 3, 5 | 2 | 100 | 90.6977 | 0.00072 | 6 |
| All | 3 | 98.6047 | 97.6744 | 0.02335 | 9 |
| 1, 2, 3, 5 | 3 | 100 | 91.8605 | 9.374e-11 | 9 |
| All | 4 | 98.6047 | 97.6744 | 0.02335 | 12 |
| 1, 2, 3, 5 | 4 | 100 | 97.6744 | 7.4547e-8 | 12 |

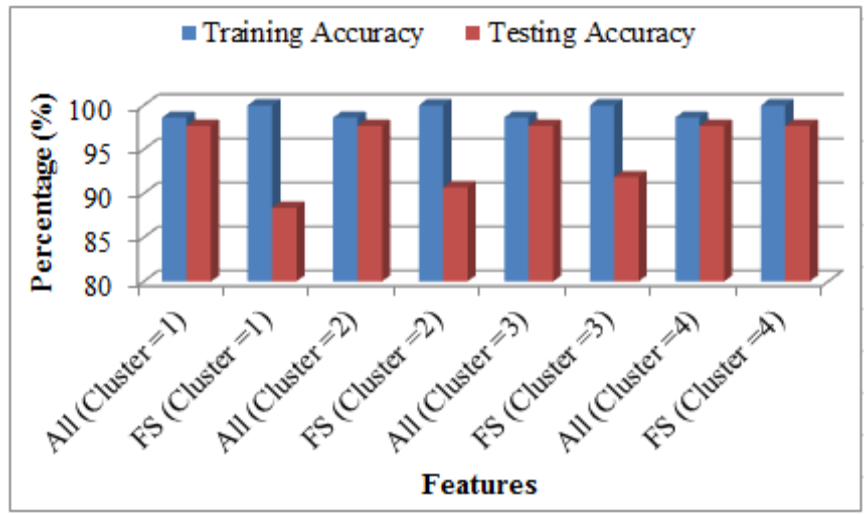

Fig. 4 LHNFCSF Classification Results of thyroid disease based on feature selection and cluster size for each class

The results indicated that, the selected features increase the recognition rate for test set. It means that some overlapping classes can be easily distinguished by selected features. The neural fuzzy classifier surface of feature 1 and feature 2 using 12 fuzzy rules is shown in Fig.

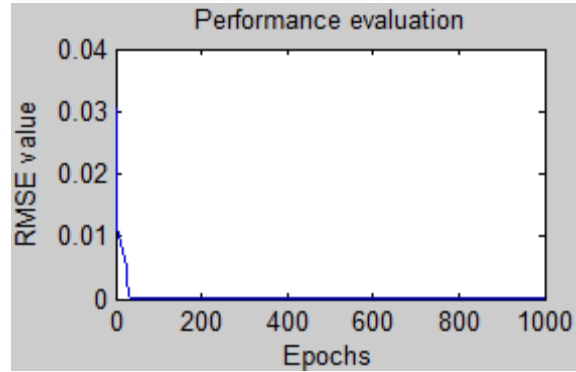

Fig. 5 Performance Evaluation of LHNFCSF during training phase after selection of relevant features using two clusters for each class

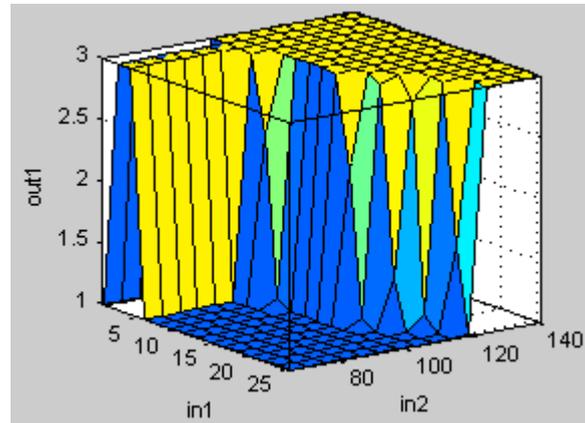

Fig. 6 Neural Fuzzy classifier surface using four clusters for each class

## 5   Conclusion

Nowadays, CAD systems are getting more and more popular. Because with the help of the CAD systems, the possible errors experts made in the course of diagnosis can be avoided, and the medical data can be examined in shorter time and more detailed as well. In fact, thyroid function diagnosis can be formulated as the classification problem, so it can be automatically performed with the aid of the CAD systems. Machine learning techniques are increasingly introduced to construct the CAD systems owing to its strong capability of extracting complex relationships in the bio- medical data. Recently, various methods have been presented to solve this problem. In this study, the positive effect of linguistic hedges on adaptive neural-fuzzy classifier is presented. According to the proposed method of Cetişli [21, 22], linguistic hedges are used in fuzzy classification rules, and adapted during the training of the system. Experimental results showed that when the linguistic hedge value of the fuzzy classification set in any feature is close to 1, this feature is relevant for that class, otherwise it may be irrelevant. The results strongly suggest that Adaptive Neuro-Fuzzy Classifier with Linguistic Hedges (ANFCLH) can aid in the diagnosis of thyroid disease and can be very helpful to the physicians for their final decision on their patients. The future investigation will pay much attention to evaluate Neuro-Fuzzy Classifier with Linguistic Hedges in other medical diagnosis problems like micro array gene selection, internet, and other data mining problems. Therefore, the impressive results may be obtained with the proposed method and improving the performance of NFCs using high-performance computing techniques. In addition, the combination of the approaches mentioned above will yield an efficient fuzzy classifier for a lot of applications.